\documentclass[conference]{IEEEtran}
\IEEEoverridecommandlockouts
\usepackage{cite}
\usepackage{amsmath,amssymb,amsfonts}
\usepackage{algorithmic}
\usepackage{graphicx}
\usepackage{textcomp}
\usepackage{xcolor}
\usepackage{caption}
\usepackage{booktabs}
\usepackage[bottom]{footmisc}
\usepackage{booktabs}

\def\BibTeX{{\rm B\kern-.05em{\sc i\kern-.025em b}\kern-.08em
    T\kern-.1667em\lower.7ex\hbox{E}\kern-.125emX}}
\begin{document}

\title{Beyond Extraction: Contextualising Tabular Data for Efficient Summarisation by Language Models\\}

\author{
\IEEEauthorblockN{Uday Allu, Biddwan Ahmed, Vishesh Tripathi}
\IEEEauthorblockA{\textit{NLP Research Team} \\
\textit{Yellow.ai}\\
\{uday, biddwan, vishesh.tripathi\}@yellow.ai}
}

\maketitle

\begin{abstract}
The conventional use of the Retrieval-Augmented Generation (RAG) architecture has proven effective
for retrieving information from diverse documents. However, challenges arise in handling complex
table queries, especially within PDF documents containing intricate tabular structures. This research
introduces an innovative approach to enhance the accuracy of complex table queries in RAG-based
systems. Our methodology involves storing PDFs in the retrieval database and extracting tabular
content separately. The extracted tables undergo a process of context enrichment, concatenating
headers with corresponding values. To ensure a comprehensive understanding of the enriched data, we
employ a fine-tuned version of the Llama-2-chat language model for summarisation within the RAG
architecture. Furthermore, we augment the tabular data with contextual sense using the ChatGPT 3.5
API through a one-shot prompt. This enriched data is then fed into the retrieval database alongside
other PDFs. Our approach aims to significantly improve the precision of complex table queries,
offering a promising solution to a longstanding challenge in information retrieval.
\end{abstract}

\vspace{10pt}


\begin{figure*}[!t]
  \centering
  \includegraphics[width=160mm]{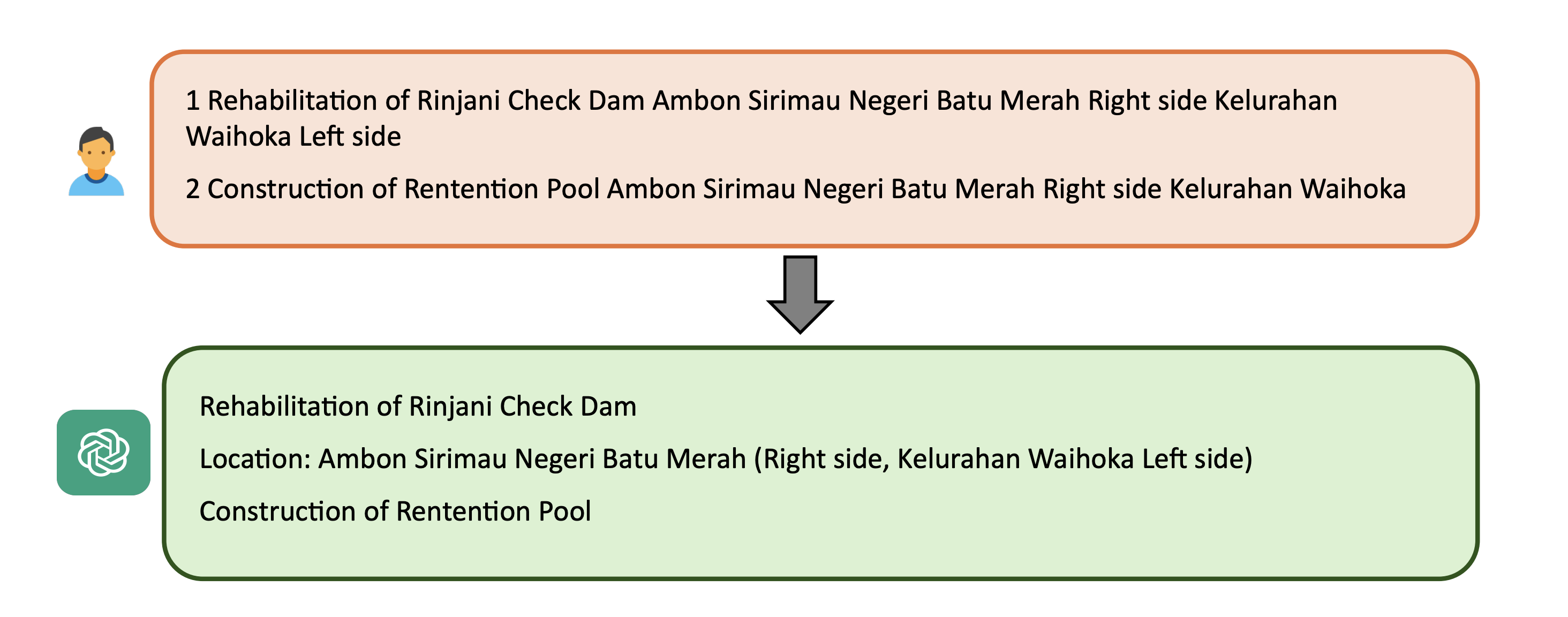}
  \captionsetup{justification=centering} % Center the caption
  \caption{The figure illustrates the Tabular Data Enhancement Assistant's task, enhancing clarity by adding headings to row values and generating coherent sentences for tabular data extracted from PDFs. It provides a guide for the assistant, highlighting the desired outcome of improving tabular content understandability.} 
  \label{fig1}
\end{figure*}

\section{Introduction}
In the era of information retrieval, the Retrieval-Augmented Generation \cite{RAG} architecture stands as a robust framework for retrieving pertinent information from diverse documents. However, its efficacy faces a substantial hurdle when confronted with complex table queries, particularly within PDF \cite{pdf_ref} documents housing intricate tabular structures. This persistent challenge has prompted the development of a novel approach aimed at elevating the accuracy of complex table queries within RAG systems. Some previous studies \cite{previous_studies} have been done, but no significant improvements have been made, especially when it comes to modifying the extracted data. This highlights the need for innovative solutions to address the limitations observed in existing methodologies

\vspace{0.2cm}

Our approach begins by addressing the inherent limitations of RAG  when dealing with tabular content. Instead of relying solely on textual retrieval, we advocate for a two-fold strategy. Firstly, PDF documents are stored in the retrieval Vector database \cite{taipalus2023vector}, ensuring a comprehensive repository of the original data. Secondly, a meticulous extraction process is implemented to separate and enrich tabular content. The enrichment process involves combining the headers and their corresponding values within the tables of PDF documents. This concatenation ensures that the context within complex rows is preserved, creating a more cohesive representation of the tabular content. By linking headers with their associated data, the augmented information becomes more structured and interpretable, allowing for improved understanding and accuracy in responding to complex table queries within the Retrieval-Augmented Generation architecture.

\vspace{0.2cm}

To imbue a deeper understanding of this enriched data, we integrate a fine-tuned version of the Llama-2-chat \cite{llama-2}, a large language model\cite{attention}, specifically tailored for summarisation, within the RAG architecture. This adaptation allows for a nuanced interpretation of the tabular content through effective summarisation. To further enhance contextual comprehension, the enriched tabular data undergoes an additional layer of augmentation using the ChatGPT 3.5 API \cite{Chat-GPT} with a one-shot prompt \cite{prompt}.

\vspace{0.2cm}

The fine-tuned Llama-2 model for summarisation ensures a specialised capability in distilling key information. The augmented data, now possessing a refined contextual sense, is seamlessly integrated into the retrieval database alongside the original PDFs. This multifaceted approach aims to significantly enhance the accuracy of complex table queries, addressing a long-standing gap in information retrieval methodologies. As we delve into the intricacies of our methodology, this paper unfolds the layers of innovation driving a paradigm shift in the domain of document-based information retrieval

\section{Methodology}

\subsection{Model Used in RAG Architecture}

Our methodological framework is centered on the innovative Retrieval-Augmented Generation (RAG) architecture, which has garnered much acclaim for its unparalleled proficiency in information retrieval applications. At the nucleus of our framework is the seamless integration with Llama-2, a cutting-edge large language model  explicitly fine-tuned for the nuanced task of summarization using massive datasets spanning various domains.The aptitude of Llama-2 in effectively condensing voluminous information into succinct summaries with remarkable coherence makes it an ideal companion for RAG architectures. During text generation, Llama-2 enhances the contextual awareness of the framework through selective retrieval of the most relevant knowledge from the provided documents and past context. The retrieved knowledge provides pertinent factual details and language cues to the generative model, guiding it to produce summaries that are abstractive yet grounded in the source content.Our methodology undertakes multi-step training of Llama-2 on large datasets to enable it to develop a comprehensive understanding of summarization across diverse topics and styles. The models are fine-tuned using supervised learning techniques that leverage gold-standard human written summaries as targets. This helps Llama-2 intrinsically build advanced capabilities for identifying and connecting key information from retrieved knowledge while generating cohesive, succinct summaries reflecting the essence of source texts.By bringing together the learned knowledge extraction strengths of Llama-2 and the informed text generation capacity of RAG, our framework is uniquely positioned to deliver superior summarization performance on complex real-world tasks, fulfilling core objectives such as generating accurate meeting notes, project reports and literature reviews.

\subsection{Dataset}

The bedrock of our investigation is a meticulously curated dataset culled from a comprehensive repository of policy documents issued by authoritative entities. This judiciously assembled corpus, designed to closely mirror the complexities and multidimensionality of real-world information retrieval, comprises a diverse array of policy domains reflecting the intricate tapestry of contemporary governance. To rigorously evaluate the Retrieval-Augmented Generation (RAG) architecture in this context, we meticulously engineered a suite of 200 targeted queries. Each query functions as a finely calibrated instrument, crafted to probe the architecture's prowess in navigating both the nuanced realm of textual information and the structured domain of tabular data. Notably, we placed a deliberate emphasis on achieving equilibrium in the representation of text-focused and table-related inquiries, ensuring a robust and well-rounded evaluation of the architecture's capabilities.

\section{Experiment Setup}

\begin{figure}[h] % You can use [h] to suggest placement "here"
\includegraphics[width=90mm]{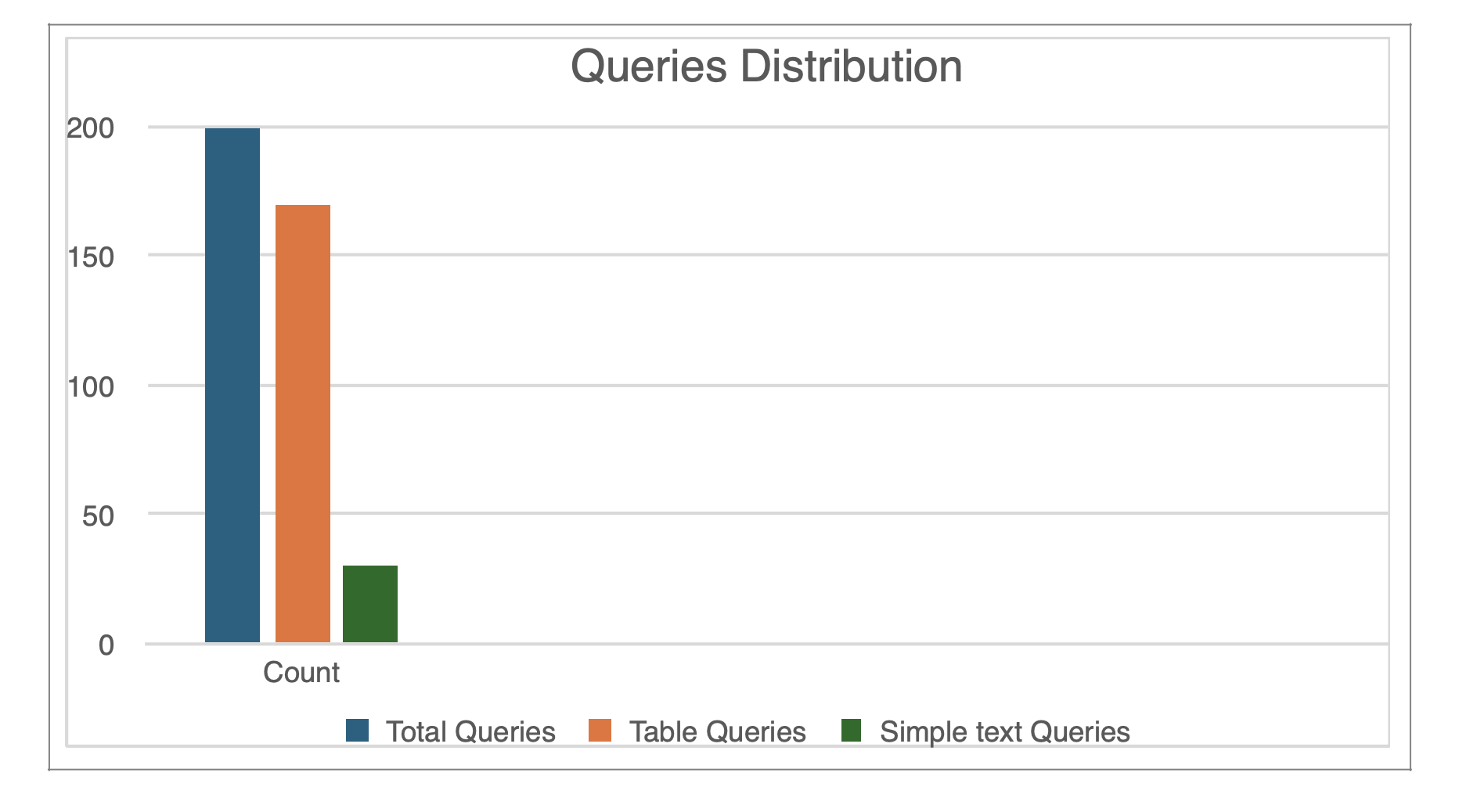}
\caption{Visualizes query distribution in the experimental dataset: 170 complex table queries and 30 simple text queries out of a total of 200. Offers a balanced evaluation of Retrieval-Augmented Generation (RAG) architecture for both textual and tabular dimensions} 
\label{fig}
\end{figure}

\begin{figure*}[!t]
  \centering
  \includegraphics[width=150mm]{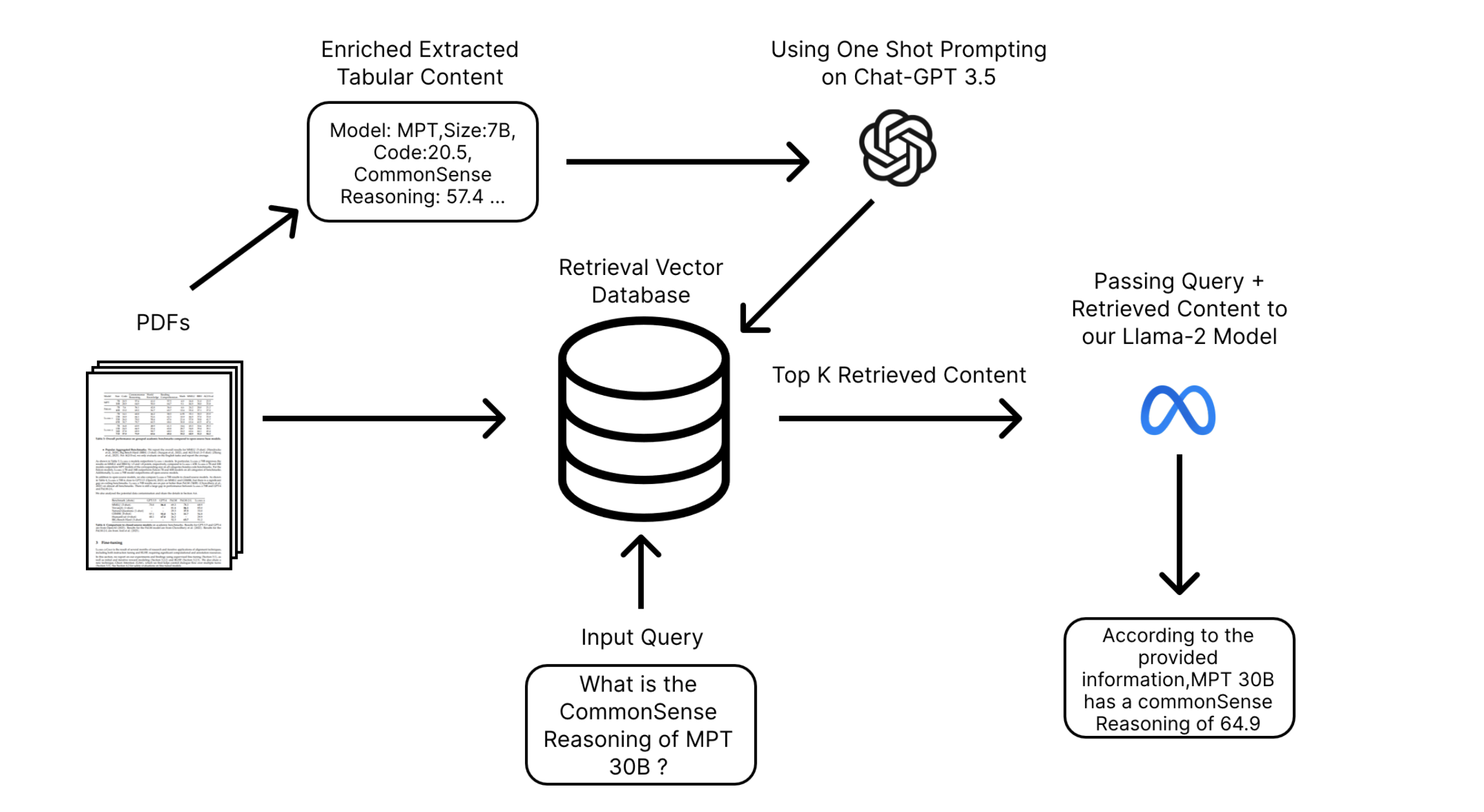}
  \captionsetup{justification=centering} % Center the caption
  \caption{Figure Represents presents the architectural diagram illustrating the experimental setup. The architecture is designed to demonstrate the workflow of our approach, showcasing the key components involved in enhancing the accuracy of complex table queries within the Retrieval-Augmented Generation (RAG) framework} 
  \label{fig}
\end{figure*}

\subsection{Query Processing}

Our experimental design meticulously orchestrated a suite of 200 diverse queries, mirroring the multifaceted nature of real-world information retrieval. This deliberate composition comprised 170 queries delving into the intricacies of tabular data. Within this domain, we further established a nuanced dichotomy: 110 complex queries probed the architecture's ability to parse intricate relationships and patterns within tables, while 60 assessed its proficiency in handling simpler tabular structures. Additionally, 30 control benchmarks, focusing on non-tabular text queries, provided a robust baseline for gauging the architecture's comprehension of unstructured text information. This comprehensive query set ensured a rigorous and well-rounded evaluation of the architecture's capabilities across a broad spectrum of information retrieval tasks.

\subsection{Data Preparation}

A pivotal facet of our methodological framework lies in the scrupulous preparation of data. Initially, PDF documents were judiciously archived within the retrieval database, creating an expansive reservoir of raw data. Subsequently, our approach veered towards a bifurcation strategy, wherein tabular content underwent a meticulous extraction process using the Camelot library, resulting in a repository of tables imbued with intricate structures. To catalyse context
enrichment, we embarked on a journey of amalgamation, deftly concatenating column headings with corresponding row values, thereby bestowing upon the extracted tables a deeper and more nuanced contextual essence

\subsection{Integration with Chat-GPT}

In the symphony of our approach, the integration of advanced language models emerges as a crescendo of innovation. The Llama-2 model, an inherent constituent of the RAG architecture, permeated the experimental landscape. Its role was not merely perfunctory; rather, it served as the discerning arbiter of summarisation tasks, illuminating the tabular data with a refined understanding. Further enhancing the contextual tapestry, the tabular data traversed an additional stratum of augmentation through the ChatGPT 3.5 API. A one-shot prompt acted as the crucible, refining the data and imbuing it with a heightened sensibility, rendering it more palatable to the discerning faculties of our summarisation model. This meticulously choreographed integration culminated in the seamless storage of the enriched data within our retrieval repository, nestled alongside the original PDFs. This archival strategy was meticulously devised to furnish our summarisation model with an augmented understanding, transcending the complexities of information retrieval and fostering an environment conducive to heightened accuracy

% \begin{figure}[h] % You can use [h] to suggest placement "here"
% \includegraphics[width=90mm]{Chart.png}
% \caption{Visualizes query distribution in the experimental dataset: 170 complex table queries and 30 simple text queries out of a total of 200. Offers a balanced evaluation of Retrieval-Augmented Generation (RAG) architecture for both textual and tabular dimensions} 
% \label{fig}
% \end{figure}

\begin{table*}[t]  % Use table* for a full-width table
\centering
\begin{tabular}{@{}cccc@{}}
\toprule
\textbf{Methodology} & \textbf{\begin{tabular}[c]{@{}c@{}}Simple Text Queries\\ Accuracy (\%)\end{tabular}} & \textbf{\begin{tabular}[c]{@{}c@{}}Table Queries\\ Accuracy (\%)\end{tabular}} & \textbf{\begin{tabular}[c]{@{}c@{}}Overall Accuracy\\ (\%)\end{tabular}} \\ \midrule
\begin{tabular}[c]{@{}c@{}}Normal Existing\\ Pipeline\end{tabular} & 86.6 & 48.2 & 54 \\ \midrule
\begin{tabular}[c]{@{}c@{}}Table Extracting\\ separately \& Context\\ Enrichment\end{tabular} & 86.6 & 54.1 & 59.4 \\ \midrule
\begin{tabular}[c]{@{}c@{}}Parsing Enriched\\ Extracted Text to ChatGPT 3.5 Turbo\end{tabular} & 93.3 & 61.1 & 66 \\ \bottomrule
\end{tabular}

\vspace{10pt}
\captionsetup{justification=centering}
\caption{Summarizes experiment outcomes, evaluating three methodologies for information retrieval accuracy. Improved metrics observed, especially in handling complex table queries. Parsing enriched text with Chat-GPT 3.5 Turbo achieves a substantial accuracy leap to 66\%, addressing challenges in complex table structures}
\label{tab:Table1}
\end{table*}

\section{Results}

This section presents the outcomes of our meticulously designed experiments, comparing the performance of three distinct policy document summarization methods on our curated dataset of 200 queries. To comprehensively evaluate the efficacy of different approaches in policy document summarization, we meticulously designed a series of experiments involving three distinct methods and a curated dataset of 200 queries. As Table-I outlines, each method was assessed on its accuracy in retrieving relevant information from both textual and tabular sections of the documents.

\subsection{Normal Existing Pipeline}

The conventional pipeline, serving as the baseline, solely extracts text from the documents and feeds it into a retrieval database. While achieving a respectable accuracy of 86.6\% for text-based queries, its performance plummets to 48.2\% for table-related queries. This stark drop highlights the inherent limitations of text-only approaches in comprehending the complexities of structured data within tables. Overall, the baseline approach yields an average accuracy of 54\%, demonstrating the significant challenges posed by integrating tabular data into the summarization process.

It's worth noting that we employed the Camelot library for our table extraction method. Despite the advanced capabilities of Camelot, the challenges associated with intricate table structures still contributed to the lower performance in table-related queries.

\subsection{Table Extracting Separately and Context Enrichment}

Our second approach introduces a dedicated stage for extracting and enriching the context of tabular data. This refined method maintains the baseline's 86.6\% accuracy for text queries, while exhibiting a dramatic improvement in table query performance, achieving an accuracy of 54.1\%. This substantial leap, resulting in an overall accuracy of 59.4\%, confirms the effectiveness of our strategy in addressing the challenges of table-based information retrieval.

\subsection{Parsing Extracted Text to Chat-GPT 3.5 API}

The pinnacle of our methodology lies in the integration of the Chat-GPT 3.5 API for parsing extracted text, unveiling a striking leap in performance. Remarkably, the accuracy for text queries surged to 93.3\%, underscoring the adeptness in handling the intricacies of unformatted text. This notable improvement is particularly noteworthy as it attests to the robustness of our approach in retrieving relevant information from less structured textual data. Simultaneously, the accuracy for table queries experienced a substantial boost, reaching 61.1\%. This transformative augmentation represents a significant stride in the adept handling of complex table queries. The amalgamation of robust text query accuracy and enhanced table query performance culminated in an impressive overall accuracy of 66\%. This achievement signifies a noteworthy advancement in our methodology’s capacity to navigate and extract valuable insights from both unstructured
text and intricate table structures

\section{Conclusions}

This study embarks on a rigorous exploration of the Retrieval-Augmented Generation (RAG) architecture's efficacy in tackling complex table queries within diverse policy documents. Through a meticulously crafted combination of model selection, dataset curation, and experimental design, we illuminate the intricacies of retrieving and comprehending information nestled within intricate tabular structures.

\vspace{0.2cm}

Our findings, meticulously outlined in the prior section, shed light on the transformative potential of our proposed methodological interventions. While traditional pipelines prove adept at handling text-based queries, they stumble when navigating the labyrinthine complexities of complex tables. Our strategy, introducing separate extraction and context enrichment for tabular data, fosters a significant upward trajectory in accuracy metrics. This nuanced approach not only bolsters the system's ability to handle table-related inquiries but also unveils the inherent challenges posed by these intricate structures.

\vspace{0.2cm}

The pinnacle of our work lies in the elegant integration of advanced language models within the RAG framework. Employing the Llama-2-chat model within the RAG architecture and subsequently parsing enriched data through the Chat-GPT 3.5 API results in a synergistic fusion that propels accuracy metrics to previously unattainable levels, particularly for complex table queries. These observed improvements highlight the profound understanding achieved through context enrichment and advanced language model integration, signaling a paradigm shift in the domain of information retrieval.

\vspace{0.2cm}

However, our journey transcends mere improvements in accuracy. It serves as a powerful testament to the boundless potential for innovation within natural language processing and information retrieval. As we meticulously navigate the landscape of our findings, it becomes readily apparent that this proposed approach not only augments the current state of the art but also lays the groundwork for future advancements in information retrieval systems. By shedding light on the efficacy of context-aware language models and RAG architectures in tackling complex table queries, we pave the way for a future where information retrieval systems effectively bridge the gap between human cognition and the intricate structures of data.

\bibliographystyle{IEEEtran}
\bibliography{library}

% \begin{thebibliography}{00}
% \bibitem{b1} P. Lewis et al., ‘Retrieval-Augmented Generation for Knowledge-Intensive NLP Tasks’, in Advances in Neural Information Processing Systems, 2020, vol. 33, pp. 9459–9474.
% \bibitem{b2} H. Touvron et al., ‘Llama 2: Open Foundation and Fine-Tuned Chat Models’, arXiv [cs.CL]. 2023.
% \bibitem{b3} M. Hardy, L. M. Masinter, D. Markovic, D. Johnson, and M. Bailey, ‘The application/pdf Media Type’, no. 8118. RFC Editor, Mar-2017.
% \bibitem{b4} T. Brown et al., ‘Language Models are Few-Shot Learners’, in Advances in Neural Information Processing Systems, 2020, vol. 33, pp. 1877–1901.
% \bibitem{b5} R. Stata, K. Bharat, and F. Maghoul, ‘The Term Vector Database: fast access to indexing terms for Web pages’, Computer Networks, vol. 33, no. 1, pp. 247–255, 2000..
% \bibitem{b6} Z. Jiang, F. F. Xu, J. Araki, and G. Neubig, ‘How Can We Know What Language Models Know?’, Transactions of the Association for Computational Linguistics, vol. 8, pp. 423–438, 07 2020.
% \bibitem{b7} F. Pan, M. Canim, M. Glass, A. Gliozzo, and J. Hendler, ‘End-to-End Table Question Answering via Retrieval-Augmented Generation’, arXiv [cs.CL]. 2022.
% \bibitem{b8} A. Vaswani et al., ‘Attention is All you Need’, in Advances in Neural Information Processing Systems, 2017, vol. 30.

% \end{thebibliography}

\end{document}